\documentclass[conference]{IEEEtran}
\IEEEoverridecommandlockouts
\usepackage{subfigure} 
\usepackage{cite}
\usepackage{amsmath,amssymb,amsfonts,amsfonts}
\usepackage{algorithmic}
\usepackage{graphicx}
\usepackage{textcomp}
\usepackage{xcolor}
\usepackage{mathrsfs}
\usepackage{booktabs}
\usepackage{balance}

\usepackage{etoolbox}
\makeatletter
\patchcmd{\@makecaption}
{\scshape}
{}
{}
{}
\makeatother
\def\tablename{Table}

\def\BibTeX{{\rm B\kern-.05em{\sc i\kern-.025em b}\kern-.08em
    T\kern-.1667em\lower.7ex\hbox{E}\kern-.125emX}}
    
\begin{document}

\title{Low-Cost GPS-Aided LiDAR State\\ Estimation and Map Building}
\author{Linwei Zheng$^{1,2}$, Yilong Zhu$^{1,2}$, Bohuan Xue$^{1,2}$, Ming Liu$^{1}$, Rui Fan$^{2,3}$\\
	$^{1}$Shenzhen Unity-Drive Innovation Technology Co. Ltd., Shenzhen, China.\\
	$^{2}$Robotics Institute, Hong Kong University of Science and Technology, Hong Kong SAR, China.\\
	$^{3}$Hangzhou ATG Intelligent Technology Co. Ltd., Hangzhou, China.\\
	{Emails: zhenglinwei@unity-drive.com, \{yzhubr, bxueaa, eelium, eeruifan\}@ust.hk}
	\vspace{-2em}
}

\maketitle

\begin{abstract}
Using different sensors in an autonomous vehicle (AV) can provide multiple constraints to optimize AV location estimation. In this paper, we present a low-cost GPS-assisted LiDAR state estimation system for AVs.  Firstly, we utilize LiDAR to obtain highly precise 3D geometry data. Next, we use an inertial measurement unit (IMU) to correct point cloud misalignment caused by incorrect place recognition. The estimated LiDAR odometry  and IMU measurement are then jointly optimized. 
We use a lost-cost GPS instead of a real-time kinematic (RTK) module to refine the estimated LiDAR-inertial odometry. Our low-cost GPS and LiDAR complement each other, and can provide highly accurate vehicle location information. Moreover, a low-cost GPS is much cheaper than an RTK module, which reduces the overall AV sensor cost. Our experimental results demonstrate that our proposed GPS-aided LiDAR-inertial odometry system performs very accurately. The accuracy achieved when processing a dataset collected in an industrial zone is approximately 0.14 m. 
\end{abstract}

\section{Introduction}
\label{sec.introduction}
Simultaneous localization and mapping (SLAM) is a very important function in autonomous vehicles (AVs) \cite{fan2018real}. Using AV sensor data and perception output, a SLAM module can not only estimate the location of an AV, but also build and update a 3D world map \cite{fan2019key}. A highly accurate global navigation satellite system (GNSS), e.g., a real-time kinematic (RTK) module or a differential Global Positioning System (GPS), can provide  centimeter-level AV localization accuracy \cite{ohno2004differential, capezio2005gps}. RTK modules utilize carrier-phase positioning data provided by base stations to compensate for the estimation error during transmission. However, the base station coverage can limit the available area of the RTK GPS receiver \cite{Langley1998}. Furthermore, RTK modules cannot solve the signal multipath problem and GPS signal shielding problem, which can occasionally make the GPS-based localization approach infeasible \cite{sunderhauf2012multipath}.

Although many researchers have managed to deal with the aforementioned problems, the solutions are very limited \cite{sunderhauf2012multipath}. On the other hand, LiDAR can provide accurate depth information, and therefore, is generally used for highly precise AV localization \cite{fan2019key, zhang2014loam, shan2018lego}. In practice, LiDAR-based localization methods perform very well in a static environment containing good geometry features. However, these methods can easily fail when there exists a dynamic object, e.g., a truck or a pedestrian. In an open environment, it is very difficult to find geometrical features when  using only LiDAR; however, GPS can work consistently. Conversely, for an environment with many buildings, LiDAR can find useful features for point cloud, but GPS will lose signal \cite{fan2019key}. In addition to GPS and LiDAR, inertial measurement units (IMUs) are generally used for ego-motion estimation \cite{fan2019key}, yet it can suffer greatly from dual integration error accumulation. The fusion of data collected by GPS, LiDAR and IMU will compensate the defects of each other. Therefore, multi-sensor systems are commonly used in autonomous cars.

\begin{figure*}[t!]
	\centering
	\includegraphics[width=0.84\textwidth]{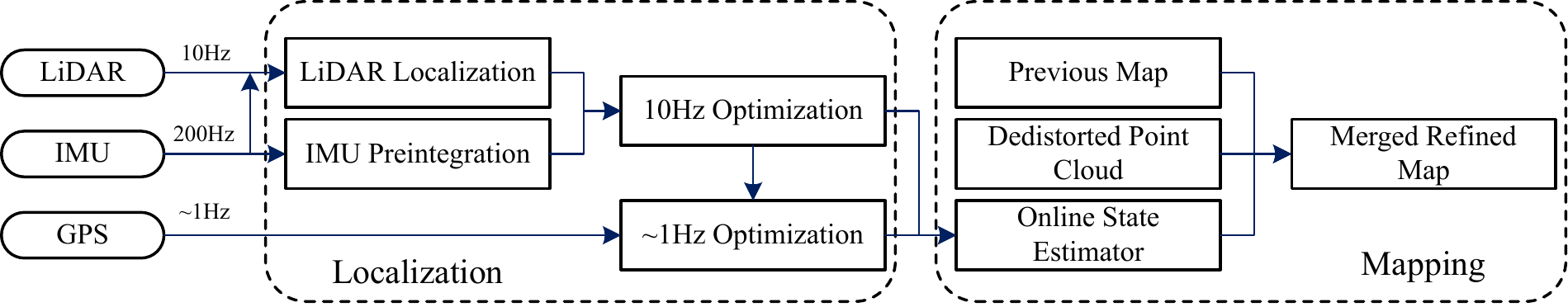}
	\caption{System pepeline}
	\label{1}
	\vspace{-1.5em}
\end{figure*}
\begin{figure}[t!]
	\centering
	\subfigure[]{
		\includegraphics[width=0.35\textwidth]{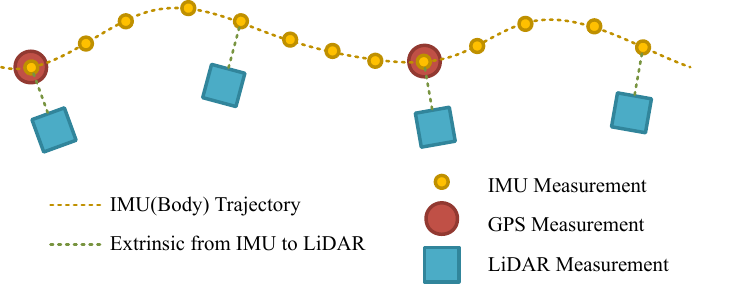}
		\label{2a}
	}
	\subfigure[]{
		\includegraphics[width=0.35\textwidth]{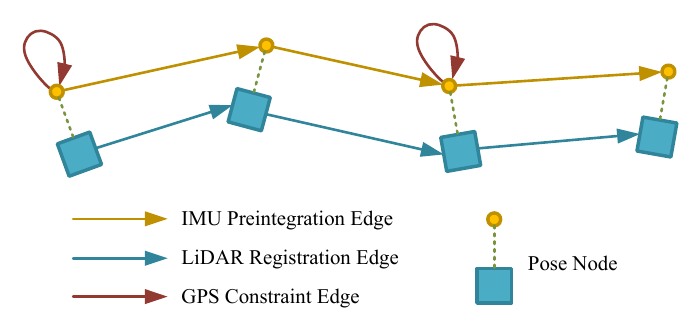}
		\label{2b}
	}
	\caption{Pose graph illustration. }
	\label{2}
		\vspace{-1.5em}
\end{figure}

\subsection{System Overview}
In this paper, we present a novel and robust LiDAR, IMU, and GPS state estimator for an AV. The system pipeline is shown in Fig. \ref{1}. The input of the system is the raw point cloud data collected using LiDAR. The point cloud is then processed to extract useful features. Next, the extracted features are used to register two different point clouds. A high-frequency IMU is used to give an initial guess for LiDAR registration. The preintegration of the IMU is considered as an edge constraint and is optimized together with the LiDAR odometry constraint. The LiDAR-inertial module can produce an optimized odometry at 10 Hz. The odometry is further optimized by integrating the GPS data at about 1 Hz. Finally, AV locations can be accurately estimated. The environment map is also built with undistorted point clouds and merged into the previous map based on loop closure detection.

In the optimization process, we consider both the sensor measurement as well as the extrinsic parameters. The proposed pose graph is shown in Fig.~\ref{2}. Between each pair of LiDAR frames, the IMU output is used to obtain an initial guess for point cloud registration. The measurement is pre-integrated and added to a pose graph \cite{kummerle2011g}, and then non-linear graph optimization is applied to refine the vehicle state. The GPS measurement is also added to the pose graph to establish an additional constraint. Based on accurate state estimation, we can build a precise environment map. To deal with the LiDAR distortion caused by movement, we introduce a continuous-time spline to de-distort each frame of the point cloud from the LiDAR. \cite{de1978practical, kim1995general, qin2000general} give us the ability to interpolate the quaternion, and the continuous-time odometry spline technique is given in \cite{lovegrove2013spline} and \cite{mueggler2018continuous}. For the reusability of the 3D map, we also merge the map on the pose graph with the method introduced in \cite{bonanni20173}. The odometry is mapped to the local east, north, up (ENU) coordinate of the start point. We tested our system in a variety of environments. We use Ceres Solver\cite{ceres-solver} to solve the non-linear least square problems in our program.

\subsection{Contributions}
The contributions of this work include:
\begin{itemize}
	\item A low-cost odometry and mapping system where the GPS constraint edge is integrated into the pose graph to optimize the LiDAR-inertial odometry. 
	\item A practical mapping method that can merge a pre-built map and newly added map, and refine each frame of the point cloud to provide a more accurate reusable map.
	\item Two datasets, one collected in an industrial zone and the other a harbor.
\end{itemize}

\subsection{Paper Structure}
The remainder of this paper is structured as follows: Section \ref{sec.related_work} discusses the related work. Section \ref{sec.li_localization} presents our LiDAR-inertial odometry method. Section \ref{sec.gps_optimization} introduces our proposed GPS optimization approach. Section \ref{sec.3d_continous_mapping} presents our 3D continuous mapping algorithm. The experimental results are illustrated and the system performance is discussed in Section \ref{sec.experimental_results}. Finally, Section \ref{sec.conclusion_future_work} concludes the paper and provides recommendations for future work.

\section{Related Work}
\label{sec.related_work}
Over the past two decades, many researchers have utilized GPS to achieve accurate odometry results \cite{ohno2004differential, capezio2005gps}. However, as discussed in Section \ref{sec.introduction}, the GPS signal is vulnerable to environment changes. Therefore, researchers have been working on environment perception and AV state estimation with the use of multiple sensors \cite{mur2015orb, qin2018vins, zhang2014loam, shan2018lego, moosmann2009segmentation}. For example, 
passive sensors such as cameras can be used to capture 2D/3D visual data of the environment \cite{fan2019roaddamage, fan2020tip}. In \cite{mur2015orb}, a vision-based odometry system named ORB-SLAM is presented, where the image feature points are used to register different input frames. However, the robustness of such a system can be easily effected by various environment conditions, e.g., light changes \cite{qin2018vins}. In addition to cameras, LiDAR, which measures the distance to an object by detecting and analyzing the reflected light \cite{fan2019key}, is also generally used for AV location estimation and 3D world map creation. 
 Compared to cameras, LiDAR is less effected by light changes, and therefore is more feasible for all-weather use \cite{fan2019key}. For instance, \cite{zhang2014loam} and \cite{shan2018lego} designed a LiDAR-based odometry system, which uses edge and surface information to register point clouds.  
 To better constrain the matching relation of the point cloud, we typically segment the ground planes in advance. For example, \cite{moosmann2009segmentation} and \cite{zhang2010lidar,chu2017fast , himmelsbach_fast_2010, marcon_dos_santos_fast_2016} present a variety of methods for point cloud segmentation. Although such methods have achieved some impressive experimental results in most situations, there still exist some drawbacks. For example, the features in the point cloud are not as rich as those extracted from an image \cite{chu2017fast}. Furthermore, especially in a wide feature-less area, a LiDAR-based system can easily fail \cite{fan2019key}.

To better estimate the sensor state, many researchers have turned their focus towards multi-sensor data fusion. EKF and UKF are two popular traditional fusion techniques demonstrated by\cite{ohno2004differential, capezio2005gps, crassidis2006sigma}. Recently, several non-linear optimization methods have been proposed to fuse different types of data.  For example, K{\"u}mmerle et al. \cite{kummerle2011g} proposed a graph-based optimization approach, which treats the robot state as a node and the constraint gained from measurement as an edge.
 Wang et al.\cite{wan2018robust} proposed a method to fuse the data collected by a multiple-sensor system that consists of  RTK modules, a LiDAR sensor, and IMUs. However, the performance of such a system is subject to GPS single point positioning.  Koide et al. \cite{koide2019portable} used an offline SLAM method based on both LiDAR and GPS.  However, the system cannot generate odometry in real time, limiting its practical application. Moreover, the LiDAR component is based on a normal distributions transform (NDT), which is not elegant. In this paper, we use the aforementioned pose graph approach but simplify the non-linear optimization problem. Furthermore,  we improve the method by utilizing a low-cost GPS instead of an RTK module, and apply feature-based point cloud registration instead of NDT. To build a clear and reusable map, we deal with the distortion of the point cloud from multi-beam LiDAR caused by the AV movement. Although \cite{zhang2014loam} has already solved this problem, the solution is based on basic linear interpolation between Euler angles. Due to the gimbal lock, the interpolation cannot always work well. Our method applies B-spline interpolation in a $\mathbb{SE}3$ space and  achieves a better de-distortion result.

\section{LiDAR-Inertial Localization}
\label{sec.li_localization}
\subsection{IMU prediction} 
We briefly denote frame as follows. $(\cdot)^W$ denotes world frame $W$, $(\cdot)^I$ denotes IMU frame $I$, and $\boldsymbol{T}^W_I$ is the IMU pose in the world frame.
In our system, the IMU state is treated as the vehicle body state. The raw measurement data includes acceleration $\boldsymbol{a}^I(t)$ and angular velocity $\boldsymbol{\omega}^I(t)$ in IMU frame $I$. $\boldsymbol{g}^W$ represents a gravity constant in the world frame $W$. Considering the measurement noise $\boldsymbol{n}$ and bias $\boldsymbol{b}$, the IMU state is as follows:
\begin{equation}
\begin{aligned}
&{\boldsymbol{\hat{\omega}}}^W= {{\boldsymbol{\omega}}}^I-{\boldsymbol{b_\omega}}^I - {\boldsymbol{n_\omega}}^I,\\
&\boldsymbol{\hat{a}}^W= \boldsymbol{R}^W_I({\boldsymbol{a}}^I -  {\boldsymbol{b_a}}^I-{\boldsymbol{n_a}}^I)+\boldsymbol{g}^W,
\end{aligned}
\end{equation}
where $\boldsymbol{R}^W_I$ is the rotation matrix from the IMU frame to the world coordinate system. We also define the translation from the IMU frame to  the world coordinate system as $t^W_I$. Then, the transformation matrix is written as $\boldsymbol{T}^W_I = [\boldsymbol{R}^W_I| \boldsymbol{t}^W_I]$. Let the IMU frame $I$ represent the state of the vehicle. We can use $\boldsymbol{T}^W_I$ to denote the pose of the vehicle in the world frame.

To predict the initial guess of the next incoming LiDAR frame, we need to integrate the IMU state during the interval between $i$ and $j$:
\begin{equation}
\begin{aligned}
&\boldsymbol{v}_j - \boldsymbol{v}_i={\boldsymbol{R}^W_I}_i ({\boldsymbol{a}}^I - {\boldsymbol{b_a}}^I- {\boldsymbol{n_a}}^I) \Delta t +{\boldsymbol{g}}^W\Delta t,
\\
& \boldsymbol{t}_j-\boldsymbol{t}_i=\boldsymbol{v}_i \Delta t + \frac{1}{2}{\boldsymbol{R}^W_I}_i  (\boldsymbol{a}^I - \boldsymbol{b_a}^I- \boldsymbol{n_a}^I) \Delta t^2+\frac{1}{2} \boldsymbol{g}^W\Delta t^2,\\
&{\boldsymbol{R}^W_I}_i^T{\boldsymbol{R}^W_I}_j= \exp(\boldsymbol{\omega}^I-\boldsymbol{b_\omega}^I - \boldsymbol{n_\omega}^I\Delta t).
\end{aligned}
\end{equation}

\subsection{IMU Preintegration}

To avoid repeatedly integrating the measurement when solving the non-linear optimization problem, we adopt the IMU preintegration method stated in \cite{forster2015imu} and \cite{forster2016manifold} into our system. The key idea is to define a relative motion of the IMU regardless of the pose and velocity at a specific time. The preintegration model between state $i$ and state $j$ is as follows:
\begin{equation}
\begin{aligned}
&\Delta \boldsymbol{R}_{ij}=\boldsymbol{R}_i^T \boldsymbol{R}_j,
\\
&\Delta \boldsymbol{v}_{ij}= \boldsymbol{R}_i^T(\boldsymbol{v}_j - \boldsymbol{v}_i - \boldsymbol{g}\Delta t_{ij}),
\\
&\Delta \boldsymbol{t}_{ij}= \boldsymbol{R}_i^T(\boldsymbol{t}_j - \boldsymbol{t}_i - \boldsymbol{v}_i \Delta t_{ij} - \frac{1}{2}\boldsymbol{g} \Delta t_{ij}^2).
\end{aligned}
\end{equation}
$\Delta \boldsymbol{R}_{ij},\Delta \boldsymbol{v}_{ij}$, and $\Delta \boldsymbol{t}_{ij}$ are defined as the preintegrated measurement of the rotation, velocity, and translation, respectively. All symbols here are denoted in the frame $W$, so we drop superscript in the equation for brevity. Noting that the measurements are independent to state $i$ and state $j$, we do not need to recalculate the preintegrated measurement when optimizing the state. The preintegrated measurement is the constraint edge in the pose graph. We also estimate the bias from the optimization and update the bias to get a more accurate result.

\subsection{LiDAR Localization}

We use multi-beam LiDAR to obtain a 3D environment point cloud. 
There are many algorithms dealing with  point cloud registration; iterative closest point (ICP) and NDT are the famous algorithms among them. However, these algorithms cannot perform in real time. To fully exploit the ground vehicle characteristics, we first divide the point cloud according to whether or not the point belongs to the ground plane. The road plane is segmented using the approach for road curb detection \cite{zhang2010lidar}. Second, we apply the edge-surface feature-based registration \cite{zhang2014loam}. The local convex is defined to calculate the local smoothness on each beam of the LiDAR points. A larger value means that the local surface is sharper. The edge and surface are selected by the criterion of the local convex. The point cloud matching is a process of minimizing the point-line distance $\boldsymbol{d}_{\mathcal{L}}$ and point-plane distance $\boldsymbol{d}_{\mathcal{P}}$. The problem is stated as follows:
\begin{equation}
\boldsymbol{T} = \arg\underset{\boldsymbol{T}}{\min} \sum_{i\in\mathscr{E}} \boldsymbol{d}_{\mathcal{L}i} + \sum_{i\in\mathscr{S}} \boldsymbol{d}_{\mathcal{P}i} + \sum_{i\in\mathscr{G}} \boldsymbol{d}_{\mathcal{P}i},
\end{equation}
where $\mathscr{E},\mathscr{S},\mathscr{G}$ represent the sets of edge points, surface points, and ground points, respectively, and $T$ is the pose of the vehicle, also known as the transformation from the body frame $\boldsymbol{B}$ to world frame $\boldsymbol{W}$. The pose $\boldsymbol{T}$ can be calculated by solving the non-linear least square problem above. The LiDAR constraint $T$ and pre-integrated measurement $\Delta \boldsymbol{R}_{ij},\Delta \boldsymbol{v}_{ij},\Delta \boldsymbol{t}_{ij}$ are coupled to get the optimized odometry of the LiDAR-inertial system output.

\section{GPS Optimization}
\label{sec.gps_optimization}
\subsection{GPS Conversion}
The obtained GPS data is latitude, longitude, and altitude (LLA). The measurement needs to be converted to the local coordinate before optimization. The conversion consists of two steps: from LLA to Earth-centered Earth-fixed (ECEF), and from ECEF to local ENU. The conversion from LLA to ECEF is as follows:
\begin{equation}
\begin{aligned}
 &\boldsymbol{x}= (\boldsymbol{R}+\boldsymbol{h})\cos \boldsymbol{\phi} \cos \boldsymbol{\lambda}\\
 &\boldsymbol{y}= (\boldsymbol{R} + \boldsymbol{h} \cos \boldsymbol{\phi} \sin \boldsymbol{\lambda}\\
 &\boldsymbol{z}= (\frac{\boldsymbol{b}^2}{\boldsymbol{a}^2}\boldsymbol{R} + \boldsymbol{h}) \sin \boldsymbol{\phi},
\end{aligned}
\end{equation}
where $\boldsymbol{\phi}$ denotes latitude, $\boldsymbol{\lambda}$ represents longitude, $\boldsymbol{h}$ denotes altitude, and $R$ is the Earth's radius defined as:
\begin{equation}
\boldsymbol{R} = \frac{\boldsymbol{a}^2}{\sqrt{(\boldsymbol{a} \cos \boldsymbol{\phi})^2 + (\boldsymbol{b}\sin \boldsymbol{\phi})^2}},
\end{equation}
where $\boldsymbol{a}$ and $\boldsymbol{b}$ is the equatorial radius and polar radius, respectively.

The origin of the local ENU coordinate is the position of the start point. The conversion from ECEF to ENU is as follows:
\begin{equation}
\left[
\begin{matrix}
\boldsymbol{e}\\
\boldsymbol{n}\\
\boldsymbol{u}
\end{matrix}
\right]
=
\left[
\begin{matrix}
-\sin \boldsymbol{\lambda} & \cos \boldsymbol{\lambda} & 0\\
-\cos \boldsymbol{\lambda} \sin \boldsymbol{\phi} & -\sin \boldsymbol{\lambda} \sin \boldsymbol{\phi} & \cos \boldsymbol{\lambda}\\
\cos \boldsymbol{\lambda} \cos \boldsymbol{\phi} & \sin \boldsymbol{\lambda} \cos \boldsymbol{\phi} & \sin \boldsymbol{\phi}
\end{matrix}
\right]
\left[
\begin{matrix}
\boldsymbol{x}-\boldsymbol{x}_0\\
\boldsymbol{y}-\boldsymbol{y}_0\\
\boldsymbol{z}-\boldsymbol{z}_0
\end{matrix}
\right].
	\vspace{-1em}
\end{equation}
\subsection{GPS Constraint}
Since low-cost GPS does not provide a heading message, we only have knowledge of the position of the GPS frame $G$. To eliminate the influence of the obscure rotation, we install the GPS antenna on the top of the IMU.  $\mathscr{P}$ represents the set of indexes of the synchronized pairs of the GPS position and IMU pose, $\boldsymbol{t}$ denotes the translation component in $[\boldsymbol{R}|\boldsymbol{t}]$, and $\boldsymbol{p}$ represents the position in the local ENU coordinate. We aim at minimizing the following residual function:
\begin{equation}
\sum_{i\in \mathscr{P}} || \boldsymbol t_i - \boldsymbol{p_i}||^2.
\end{equation}
As shown in Fig.~\ref{2b}, the GPS constraint is an unary hyper-edge, which  is connected to the IMU \textit{pose} node. Here, we keep a sliding window of recent \textit{poses} to get a refined result. The optimization considers the constraint of the preintegrated IMU measurement. 

\section{3D Continuous Mapping}
\label{sec.3d_continous_mapping}
From the previous section, we can obtain a refined odometry, which is constrained by the LiDAR, an IMU, and a GPS. In this section, we undistort the point cloud and build a high-definition reusable map. 
Please note that the multi-beam LiDAR sensor is working continuously, even though we usually obtain discrete separated frames of the point cloud from the LiDAR driver. In other words, the points with different azimuths have different time stamps. Thus, we need to undistort each frame of the point cloud to get the original scan result. For instance, the timestamp of the beginning point and ending point in a single LiDAR scan have a difference of the period $\Delta t$.

\begin{figure}[t!]
	\centering
	
	\includegraphics[width=0.33\textwidth]{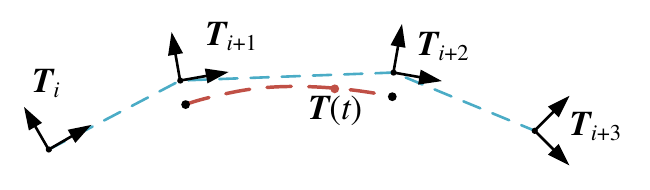}
	
	\caption{Cubic B-Spline in $\mathbb{SE}3$}
	\label{3}
	\vspace{-1em}
\end{figure}

To undistort each point cloud frame, we need to spline between the discrete odometry. We apply the uniform cubic B-Splines in $\mathbb{SE}3$ according to \cite{lovegrove2013spline} and \cite{mueggler2018continuous}. As can be seen in Fig.~\ref{3}, we need four transformation matrices $\boldsymbol{T}_i, \boldsymbol{T}_{i+1}, \boldsymbol{T}_{i+2}$, and $\boldsymbol{T}_{i+3}$ to interpolate a smooth state $\boldsymbol{T}(t)$. We have a basis matrix $\bf{\tilde{M}}$:

\begin{equation}
\begin{aligned}
\tilde {\boldsymbol{\rm M}} 
 &= \left[ {\begin{array}{*{20}{c}}
6&5&1&0\\
0&3&3&0\\
0&{ - 3}&3&0\\
0&1&{ - 2}&1
\end{array}} \right],
\end{aligned}
\end{equation}

and the cumulative basis function in matrix form is

\begin{figure}[t!]
	\centering
	\subfigure[]{
		\includegraphics[width=0.22\textwidth]{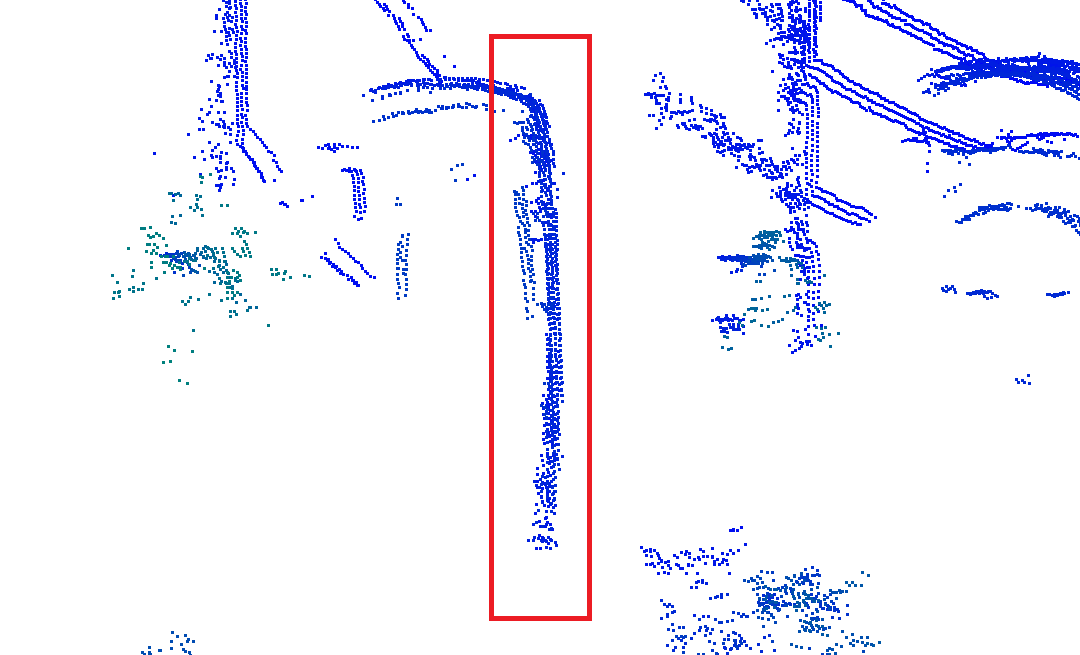}
		\label{4a}
	}
	\subfigure[]{
		\includegraphics[width=0.22\textwidth]{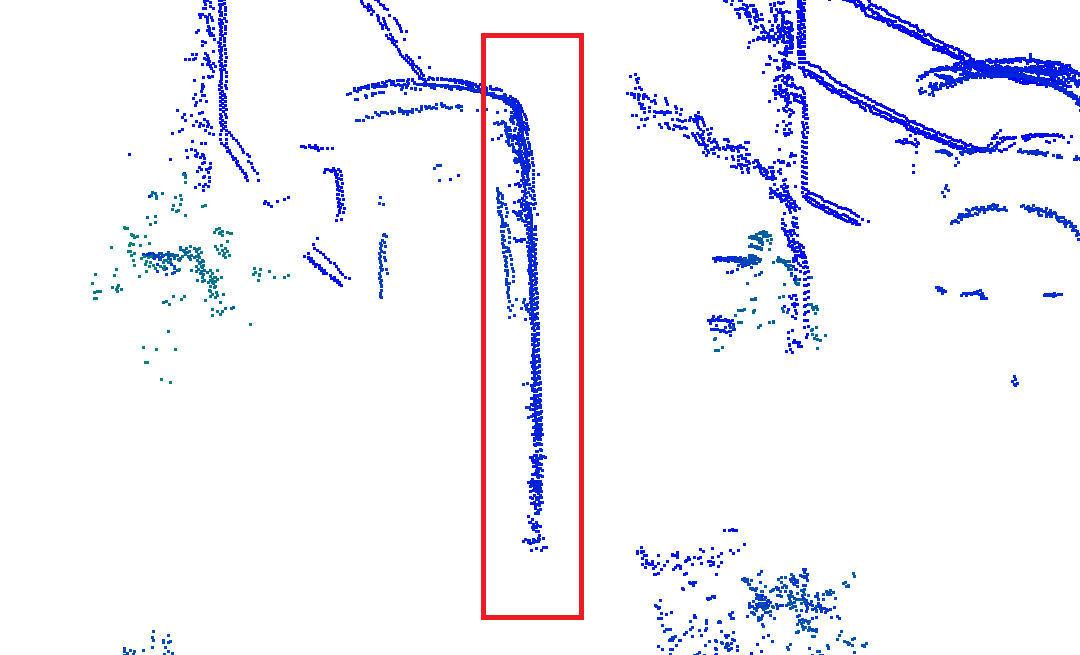}
		\label{4b}
	}
	\caption{A comparison (a) a distorted point cloud and (b) undistorted point cloud.}
	\label{4}
		\vspace{-2em}
\end{figure}
\begin{table}[t!]
	\centering
	\caption{Experimental results of the two datasets.}
	\begin{tabular}{ccccc}
		\toprule
		Dataset & max (m) & min (m) & mean (m) & SD (m)\\
		\midrule
		industrial zone & 0.9462 & 0.0061 & 0.1421 & 0.0808 \\
		Harbor & 8.7316 & 0.0027 & 2.8068 & 2.1302 \\
		\bottomrule
	\end{tabular}
	\vspace{-2em}
	\label{table1}
\end{table}

\begin{equation}
\begin{aligned}
{\bf{\tilde B}} &= \frac{1}{{3!}}\left[ {\begin{array}{*{20}{c}}
1&u&{{u^2}}&{{u^3}}
\end{array}} \right]{\bf{\tilde M}}\\
 &= \frac{1}{{3!}}{\left[ {\begin{array}{*{20}{c}}
6\\
{5 + 3u - 3{u^2} + {u^3}}\\
{1 + 3u + 3{u^2} - 2{u^3}}\\
{{u^3}}
\end{array}} \right]^{\rm{T}}},\\
u &= \frac{t-t_{i+1}}{\Delta t}, u \in [0, 1),
\end{aligned}
\end{equation}
where $\Delta t$ is the time interval between the adjacent transformations, and $t_{i+1}$ is the timestamp of the $(i+1)$-{th} transformation. We assume that the angular velocity stays constant and the acceleration is zero during the interval. Then, we can define the following equation:
\begin{equation}
\begin{aligned}
\boldsymbol{T}(t) =& \exp ({{\tilde B}_{0,4}}\log ({\boldsymbol{T}_i}))\prod\limits_{k = 1}^3 {\exp ({{\tilde B}_{k,4}}\log ({\boldsymbol{T}_{i + k - 1}}^{ - 1}{\boldsymbol{T}_{i + k}}))} \\
 =& \exp (\log ({\boldsymbol{T}_i}))\exp \left( {\frac{1}{6}(t + 3u - 3{u^2} + {u^3})\log ({\boldsymbol{T}_{ii + 1}})} \right)\\
&\exp \left( {\frac{1}{6}(1 + 3u + 3{u^2} - 2{u^3})\log ({\boldsymbol{T}_{i + 1i + 2}})} \right)\\
&\exp \left( {\frac{1}{6}{u^3}\log ({\boldsymbol{T}_{i + 2i + 3}})} \right).
\end{aligned}
\end{equation}
The spline $\boldsymbol{T}(t)$ is used to undistort the points in each frame. As shown in Fig.~\ref{4}, the original point cloud shown in Fig.~\ref{4a} is refined after the undistortion and it becomes the point cloud shown in Fig.~\ref{4b}. Then, the undistorted point cloud is superimposed onto the map. The map point cloud and the origin of the map coordinate are saved for the purpose of reuse.
After vehicle reboot, the system will load the pose graph and point cloud map. Along with the running of the vehicle, the new \textit{nodes} and \textit{edge} will be added to the pose graph. The joint optimization will merge the new and old maps. 

\begin{figure}
	\centering
	\subfigure[]{
		\includegraphics[height=0.23\textheight]{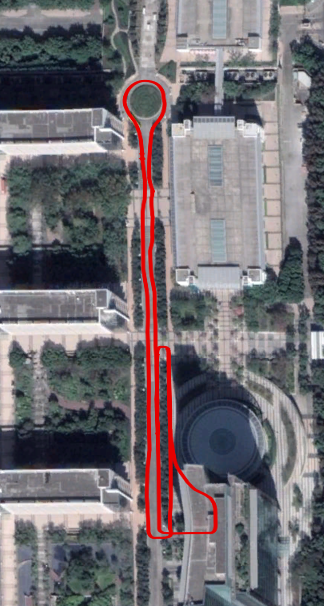}
		\label{5a}
	}
	\subfigure[]{
		\includegraphics[height=0.23\textheight]{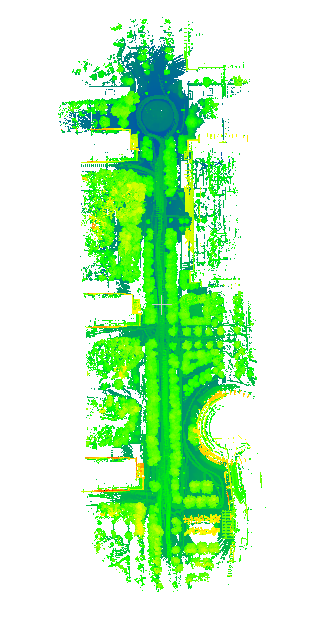}
		\label{5b}
	}
	\subfigure[]{
		\includegraphics[width=0.33\textwidth]{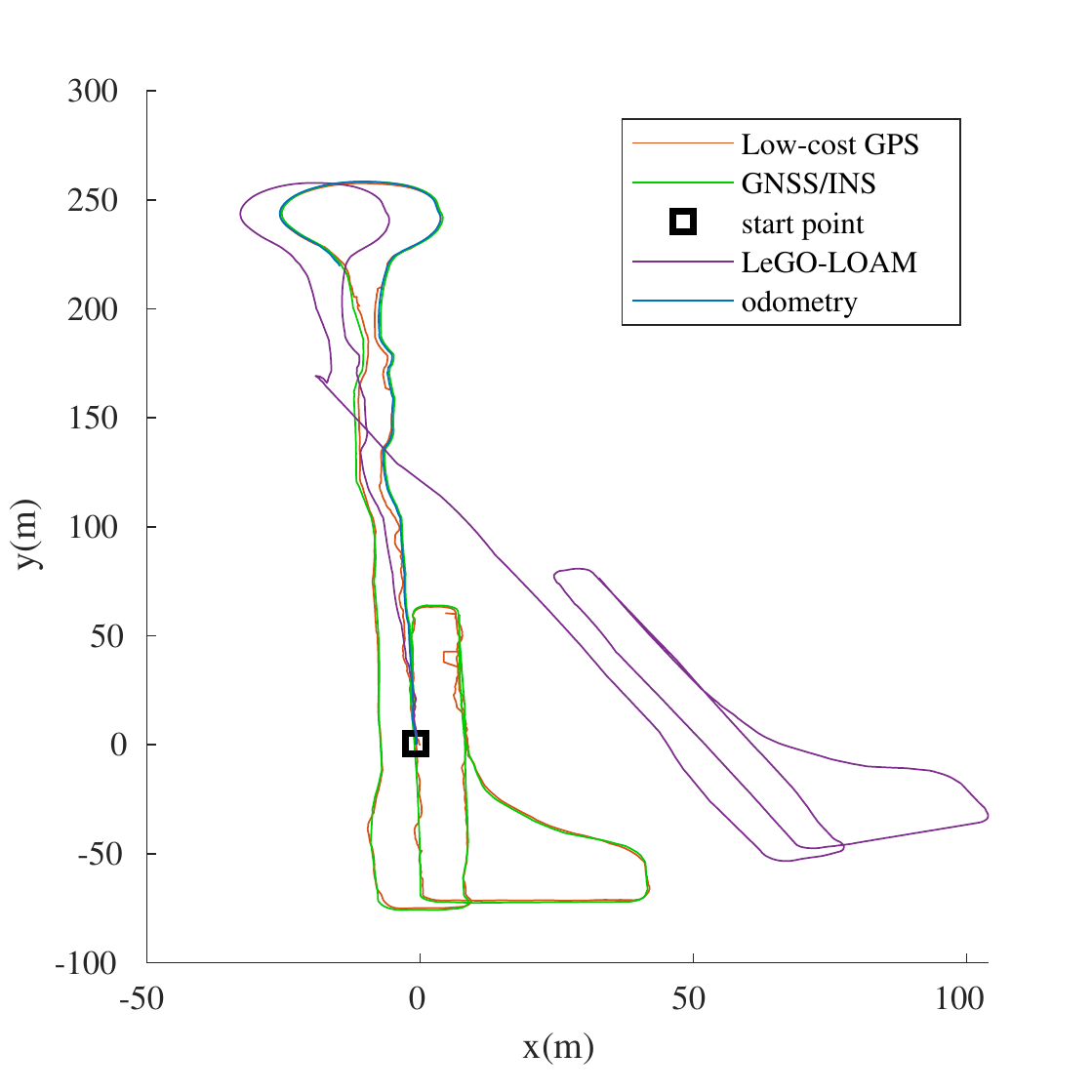}
		\label{5c}
	}
	\caption{Experimental results of the industrial zone dataset: (a) satellite view; (b) created map; (c) trajectory comparison. }
	\label{5}
	\vspace{-1.5em}
\end{figure}
\begin{figure}[t!]
	\centering
	\subfigure[]{
		\includegraphics[height=0.23\textheight]{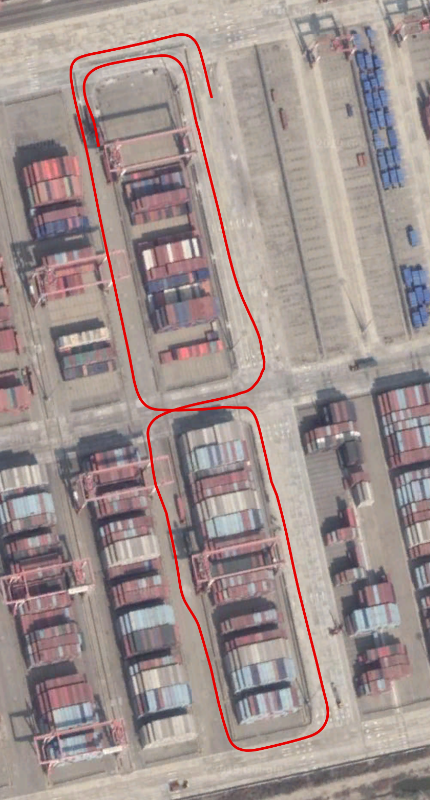}
		\label{6a}
	}
	\subfigure[]{
		\includegraphics[height=0.23\textheight]{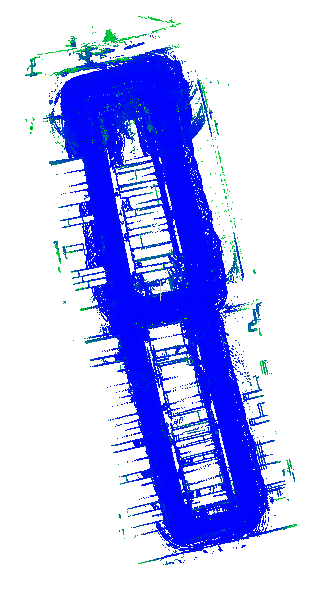}
		\label{6b}
	}
	\subfigure[]{
		\includegraphics[width=0.33\textwidth]{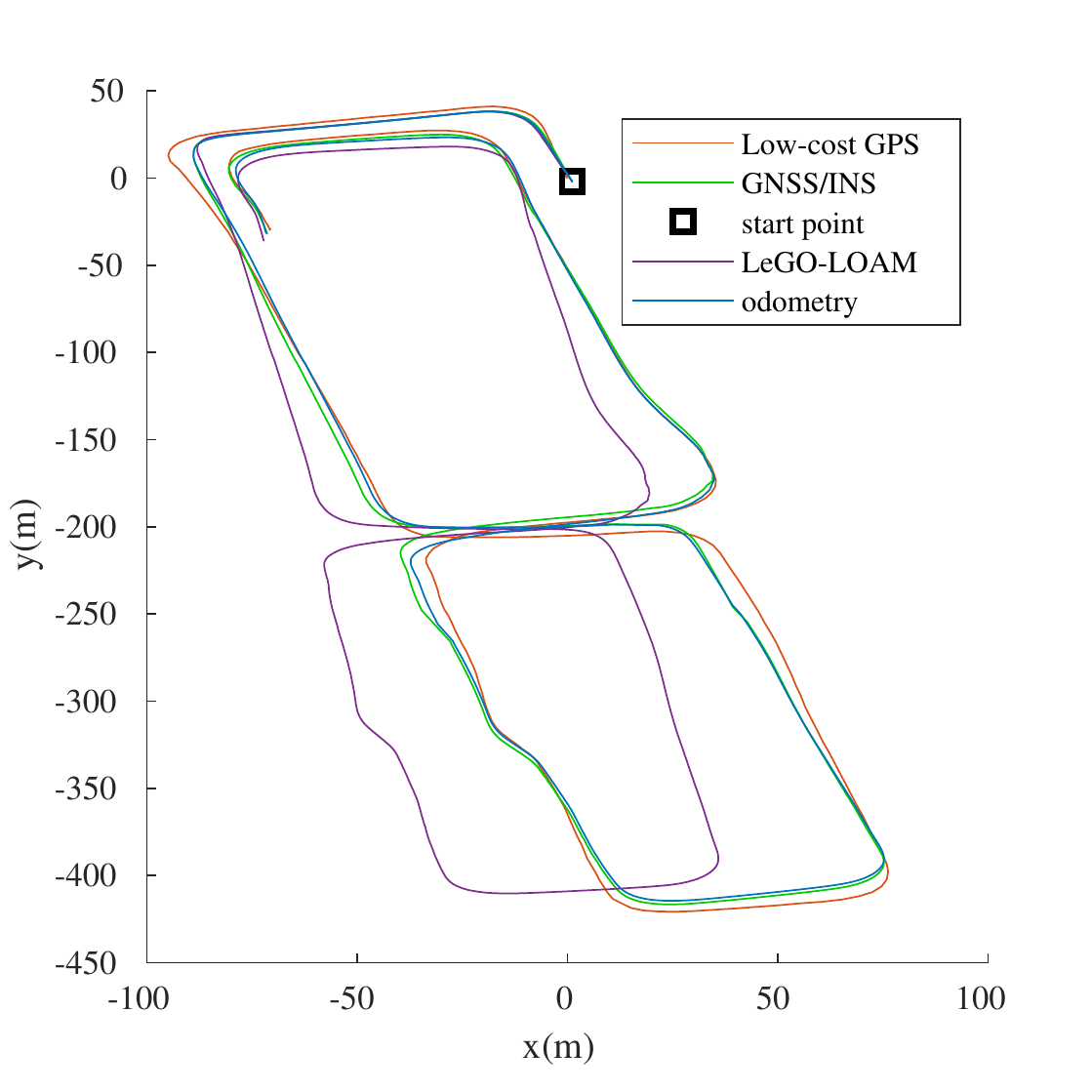}
		\label{6c}
	}
	\caption{Experimental results of the harbor dataset: (a) satellite view; (b) created map; (c) trajectory comparison.}
	\label{6}
	\vspace{-1.5em}
\end{figure}

\section{Experimental Results}
\label{sec.experimental_results}
Our vehicle was equipped with a Velodyne VLP-16 LiDAR sensor, a Ublox GYGPSV1 NEO-M8N GPS module, and an LPMS-ME1 IMU module. The LiDAR sensor has 16 channels of laser scanners and can provide a 3D point cloud at 10 Hz. The GPS module can provide data at 1 Hz, and the IMU can provide data at up to 400 Hz. We also utilized a highly precise GNSS (NovAtel 718d) to provide the ground truth with our system. The computing platform was designed based on an NUC (i7-7567U@3.50GHz, 16GB RAM) and it was assembled within the car. We implemented our algorithm in C++ under the framework of Robot Operating System (ROS). Our testing sites include an industrial zone and a wide harbor. Our experimental results are shown in Table~\ref{table1}, where SD refers to standard deviation. The next subsections detail the experimental results of the industrial zone dataset and harbor dataset, respectively.

\subsection{Industrial Zone Test}
Firstly, we tested our algorithm in an industrial zone, where the surroundings are full of trees and buildings in which GPS cannot perform well. We use a GNSS/INS to provide precise ground truth with centimeter accuracy. The total driving distance is about 1.1 km.

The experimental results are shown in Fig.~\ref{5}. The red line represents the AV trajectory provided by our low-cost GPS, and the green line represents the trajectory ground truth. Fig.~\ref{5c} shows that our method (blue) outperforms LeGO-LOAM (purple). 
 It is worthy noting that there is a huge deviation in the LeGO-LOAM trajectory at around (-20, 150). The reason is that a large truck is passing nearby at that time and LeGO-LOAM misaligns the point cloud due to the dynamic object (truck). In contrast, our method uses GPS to constrain the final odometry and provide a better trajectory, which looks very similar to the ground-truth trajectory. The mean error is only about 0.14 m.

\subsection{Harbor Test}
In addition, we also tested our algorithm in a wide harbor, where there are few obstructions in the area and our GPS works very well. In such a testing site, our LiDAR can barely find useful geometry features, such as edges and surfaces. As discussed in the previous subsection, LeGO-LOAM jumps suddenly because of the point cloud misalignment, while our method is able to trace the ground truth. This is because the GPS data plays an important role in the open area.

\section{Conclusion and Future Work}
\label{sec.conclusion_future_work}
In this paper, we presented a robust and novel state estimate system for AVs. Our approach can provide reliable real-time odometry with the support of LiDAR, an IMU, and a low-cost GPS. The proposed system also has features such as initialization, relocalization, map storage and merging, failure detection and recovery, and pose graph optimization. Our system achieved very good performance on public datasets, which were collected in a harbor and a industrial zone. Additionally, the reusable map can save many resources for online localization. The most practical contribution of the work is the use of a low-cost GPS instead of RTKs, which can also decrease the cost of an AV. 

Although our system can be used in both broad and narrow environments with the use of both GPS and LiDAR, it needs to be improved to adapt to more complex environments. Furthermore, point cloud feature extraction also needs to be refined to gain a better feature association. While the system reduces the AV sensor cost without reducing the performance, the system robustness needs further improvement. Future work will include designing a GPS-aided stereo odometry system based on our previously published disparity estimation algorithm \cite{fan2018road}.

\section*{Acknowledgment}
This work was supported by the National Natural Science Foundation of China, under grant No. U1713211, the Research Grant Council of Hong Kong SAR Government, China, under Project No. 11210017, No. 21202816, and the Shenzhen Science, Technology and Innovation Commission (SZSTI) under grant JCYJ20160428154842603, awarded to Prof. Ming Liu.
\bibliographystyle{IEEEtran}
\balance


\end{document}